\documentclass[sigconf]{acmart}
\usepackage{microtype}
\usepackage{enumitem}
\usepackage{amsmath}

\usepackage{amssymb}
\usepackage{times}
\usepackage{latexsym}
\usepackage{bbm}
\usepackage{graphicx}
\usepackage{multirow}
\usepackage{multicol}
\usepackage{bm}
\usepackage{booktabs}
\usepackage{algorithmic}
\newcommand{\change}[1]{\textcolor{black}{{#1}}}
\usepackage[linesnumbered,ruled]{algorithm2e}
\usepackage{url}
\newtheorem*{remark}{Remark}

\AtBeginDocument{%
  \providecommand\BibTeX{{%
    \normalfont B\kern-0.5em{\scshape i\kern-0.25em b}\kern-0.8em\TeX}}}

\copyrightyear{2023}
\acmYear{2023}
\setcopyright{acmlicensed}\acmConference[SIGIR '23]{Proceedings of the 46th International ACM SIGIR Conference on Research and Development in Information Retrieval}{July 23--27, 2023}{Taipei, Taiwan}
\acmBooktitle{Proceedings of the 46th International ACM SIGIR Conference on Research and Development in Information Retrieval (SIGIR '23), July 23--27, 2023, Taipei, Taiwan}
\acmPrice{15.00}
\acmDOI{10.1145/3539618.3591650}
\acmISBN{978-1-4503-9408-6/23/07}

\begin{document}

\title{\textsc{Cone}: Unsupervised Contrastive Opinion Extraction}



\author{Runcong Zhao}
\affiliation{%
  \institution{King's College London \\ University of Warwick}
  \country{United Kingdom}}
\email{runcongz@gmail}

\author{Lin Gui}
\affiliation{%
  \institution{King's College London}
  \streetaddress{Strand}
  \country{United Kingdom}}
\email{lin.1.gui@kcl.ac.uk}

\author{Yulan He}
\affiliation{%
  \institution{King's College London \\ The Alan Turing Institute}
  \streetaddress{British Library, 96 Euston Rd}
  \country{United Kingdom}}
\email{yulan.he@kcl.ac.uk}

\renewcommand{\shortauthors}{R. Zhao, L. Gui and Y. He}

\begin{abstract}
Contrastive opinion extraction aims to extract a structured summary or key points organised as positive and negative viewpoints towards a common aspect or topic. Most recent works for unsupervised key point extraction is largely built on sentence clustering or opinion summarisation based on the popularity of opinions expressed in text. 
However, these methods tend to generate aspect clusters with incoherent sentences, conflicting viewpoints, redundant aspects. 
To address these problems, we propose a novel unsupervised Contrastive OpinioN Extraction model, called \textsc{Cone}, which learns disentangled latent aspect and sentiment representations based on pseudo aspect and sentiment labels by combining contrastive learning with iterative aspect/sentiment clustering refinement. Apart from being able to extract contrastive opinions, it is also able to quantify the relative popularity of aspects and their associated sentiment distributions. The model has been evaluated on both a hotel review dataset and a Twitter dataset about COVID vaccines. The results show that despite using no label supervision or aspect-denoted seed words, \textsc{Cone} outperforms a number of competitive baselines on contrastive opinion extraction. The results of \textsc{Cone} can be used to offer a better recommendation of products and services online.
\end{abstract}




\maketitle

\section{Introduction}

\begin{figure}[t!]
    \centering
    \includegraphics[width=\linewidth]{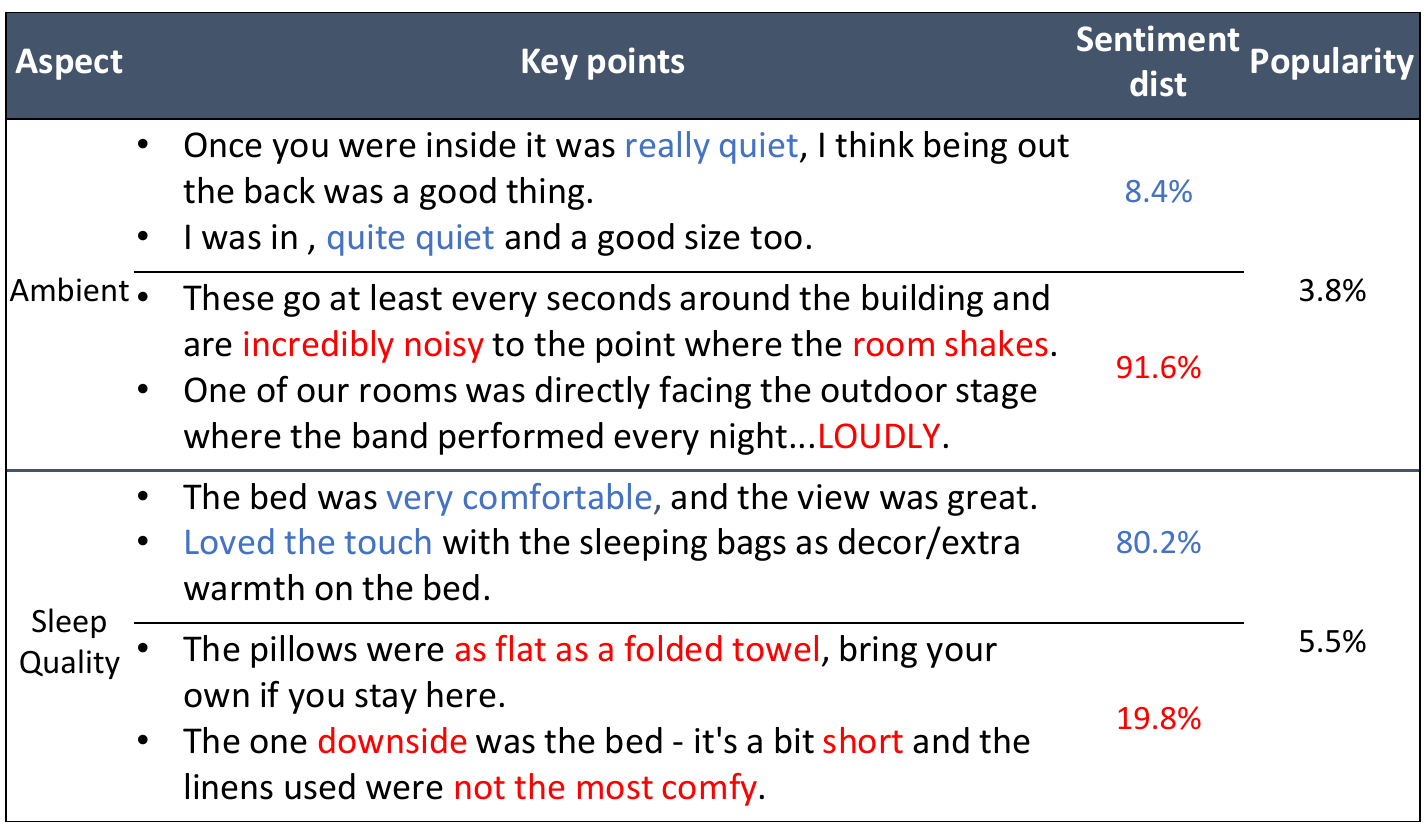}
    \caption{Example contrastive opinions extracted by our model from hotel reviews, where ``Sentiment dist.'' stands for the sentiment distribution (`\emph{\color{blue}Positive}' in {\color{blue}blue} while `\emph{\color{red}Negative}' in {\color{red}red}) under a given aspect, and ``Popularity'' stands for the occurrence proportion of the given aspect in the whole corpus.}
    \label{fig:typical-output}
\end{figure}

\change{Social media platforms produce an abundance of user-generated content. As manual summarisation is impractical, there is a need for automatic extraction of contrastive opinions expressed towards the same topic/aspect 
to understand different perspectives shared in social media. 
For example, to gain a quick glimpse of what has been discussed in hotel reviews, it would be helpful if contrastive viewpoints towards the same aspects such as such as `\emph{Ambience}' and `\emph{Sleep quality}' can be organised into coherent clusters, as shown in Figure \ref{fig:typical-output}. Moreover, the information about the relative popularity of the aspects discussed and their associated sentiment distributions would make it easier for users to cross compare multiple products in the same category (such as \emph{hotels}), and hence making more informed purchase decisions. Apart from making a better recommendation of products and services online, the results of contrastive opinion extraction can also be used in downstream tasks such as developing a better product search system since the aggregation of customer opinions at the product aspect level can be viewed as a consensus ranking problem. 
}

To identify key points discussed in text as a way for opinion extraction, it is possible to develop extractive models based on clustering \cite{2016:ptdec, topiccluster2022www} or opinion popularity \cite{kpa2020}, which does not require any labelled data for model training. 
An alternative approach  \cite{2021:qt, sum2021www, copycat2020} is through opinion summarisation which first maps input text into latent codes by a variant of vector quantization variational autoencoder and then extracts sentences which are close to the latent codes as opinion summaries based on the popularity of the opinion they express. Apart from extracting an overall opinion summary, aspect-level opinion summaries can be extracted by providing some aspect-denoting query terms. Nevertheless, the aforementioned approaches do not explicitly separate the extracted opinions by their polarities. Therefore, the resulting key points or summaries may contain conflicting viewpoints. 
Another shortcoming of existing methods 
is that most of them rely on text representations encoded using pre-trained-language models (PLMs). In the absence of supervised signals such as class labels to fine-tune PLM-encoded text representations, it is difficult to disentangle embeddings representing aspects and those representing sentiment. 
As a result, approaches based on PLM-encoded embeddings tend to produce incoherent and redundant aspect clusters, and suffer from information loss as they tend to discard edge cases  
in order to maintain the quality of the extracted opinions.

To address the limitations of existing methods, we propose a novel unsupervised \textbf{C}ontrastive \textbf{O}pinio\textbf{N} \textbf{E}xtraction approach, called \textsc{Cone}, which combines contrastive learning \cite{cert2020, SimCSE2021} with iterative aspect/sentiment clustering refinement\footnote{Our code can be found at \url{https://github.com/BLPXSPG/CONE}.}. Here, contrastive learning aims to learn disentangled latent aspect and sentiment representations under the unsupervised learning setup, while the proposed clustering refinement strategy aims to produce better aspect/sentiment clustering results which can be guaranteed by our mathematical proof. 

In practice,  for a given document, each sentence is mapped to two separate latent spaces for automated sentiment and aspect disentanglement 
through our designed \emph{denoised contrastive learning} framework. The key to contrastive learning is the construction of positive and negative training instances. The positive instance is obtained by backtranslation, that is, translating the original sentence into another language which is then translated back to English.  The negative instances are sampled from different documents with different sentiment and aspect pseudo labels. Here, we obtain the sentiment pseudo labels from a rule-based sentiment classifier, while getting the aspect pseudo labels by performing $k$-means clustering on sentence embeddings encoded by SBERT \cite{sentbert2019}. We have shown theoretically that our proposed negative sampling strategy can guarantee producing better results compared to randomly sampling sentences from different documents without considering pseudo labels, under the condition that the clustering accuracy is better than random guessing. 
The learned sentence-level latent sentiment/aspect representations are then used for clustering. Similar clusters are merged by a threshold modelled by a log-normal distribution. The new clustering results are used to update the sentiment and aspect pseudo labels, which are in turn used to adjust latent sentiment and aspect representations through contrastive learning. 
The above process is repeated until the silhouette score of the aspect clusters converges. 
We have shown empirically that contrastive learning helps generate key points with more lexical variations under the same aspect cluster, as opposing to conventional cluster-based approaches where 
the top sentences under each aspect are almost the same. As a side product, our approach is able to display the relative popularity of each aspect and its associated sentiment distribution. In summary, the contributions of our work is four-fold: 
\begin{itemize}[noitemsep]
    \item We propose a new framework, which combines contrastive learning with iterative aspect/sentiment clustering refinement, for unsupervised contrastive key point extraction. 
    \change{It is able to identify more informative aspects compared to existing unsupervised keypoint extraction approaches}. 
    \item We design a novel contrastive learning framework, \change{to address the problem of conflicting viewpoints}, 
    which learns disentangled latent aspect and sentiment representations based on positive and negative training instances constructed by our proposed sampling strategies built on pseudo aspect/sentiment labels. 
    \item We propose to model the threshold of merging similar aspect clusters as a log-normal distribution during clustering refinement \change{to alleviate the problem of redundant aspects}.
    \item Experimental results show that our proposed framework outperforms several strong baselines on contrastive opinion extraction, despite using no supervised information or aspect-denoted seed words.
\end{itemize}

\section{Related work}

Opinion extraction aims to extract topics or aspects discussed in text and their associated sentiments or attitudes from documents. 
Early approaches are largely built on the Latent Dirichlet allocation (LDA)  model \cite{melda2010,sas2012}. 
These models assume that the input text is generated from a mixture of latent aspects and each aspect has its probabilistic distribution on words. 

In recent years, extracting aspects using neural models with additional information incorporated has become more popular. Examples include the models using word co-occurrences \cite{attentionae2017}, considering dependency relations between sentences \cite{rnnae2018}, or separating embeddings into general-purpose and domain-specific ones \cite{cnnae2018}. To present the extracted opinions, the aforementioned models typically extract representative aspect words or choose sentences which are most similar to the top topic words based on some similarity measurement. In such cases, the extracted aspects might not form coherent viewpoints. 
Another branch of studies aims to use prior knowledge to guide the learning of option extraction, such as expert prior opinion \cite{contraop2015www, cikm2016lim, KoMen2022}, argument and stance categories \cite{kpace2021}, external knowledge base\cite{sumdocs2021}, aspect and rating labels \cite{coling2010lu}, and seed words in opinionated text \cite{kpa2020, aspectsent2021sigir}. 
For example, Key point analysis (KPA) \cite{kpa2020} first extracts a set of high-level statements, which are defined as the key points, and then identifies supporting evidence for each key point through semantic similarity mapping. 
KPA requires a classifier to identify high quality key points, which is trained on a small set of labelled data. 
KPA was later extended to train with gold-standard arguments and key points by utilising contrastive learning and extractive argument summarisation \cite{kpace2021}.

Apart from aspect-level key point extraction, opinion summarisation \cite{multinews2019, extrasum2020} 
can also be used to extract 
opinions from input text. However, most existing opinion summarisation approaches only generate summaries with mixed polarities 
\cite{icml2019chu}. 
Some work performs post-processing by removing sentences of the same aspect but with different polarities from the summary \cite{wsdm2022ke}. Nevertheless, these approaches are unable to provide summary statistics about the relative popularity of the aspects discussed in text and the distribution of their associated sentiments. 
Apart from an overall opinion summary, it is possible to generate the summaries at the aspect level \cite{2021:qt, 2019ACLJohn}. 
But such approaches often require a pre-defined set of aspects and a list of manually defined seed words for each aspect. 
For other types of data such as tweets about vaccine attitudes, it is impossible to pre-define \change{a comprehensive set of aspects} due to the constantly evolving topics being discussed online.

\begin{figure*}[htb]
    \centering
    \includegraphics[width=0.84\linewidth]{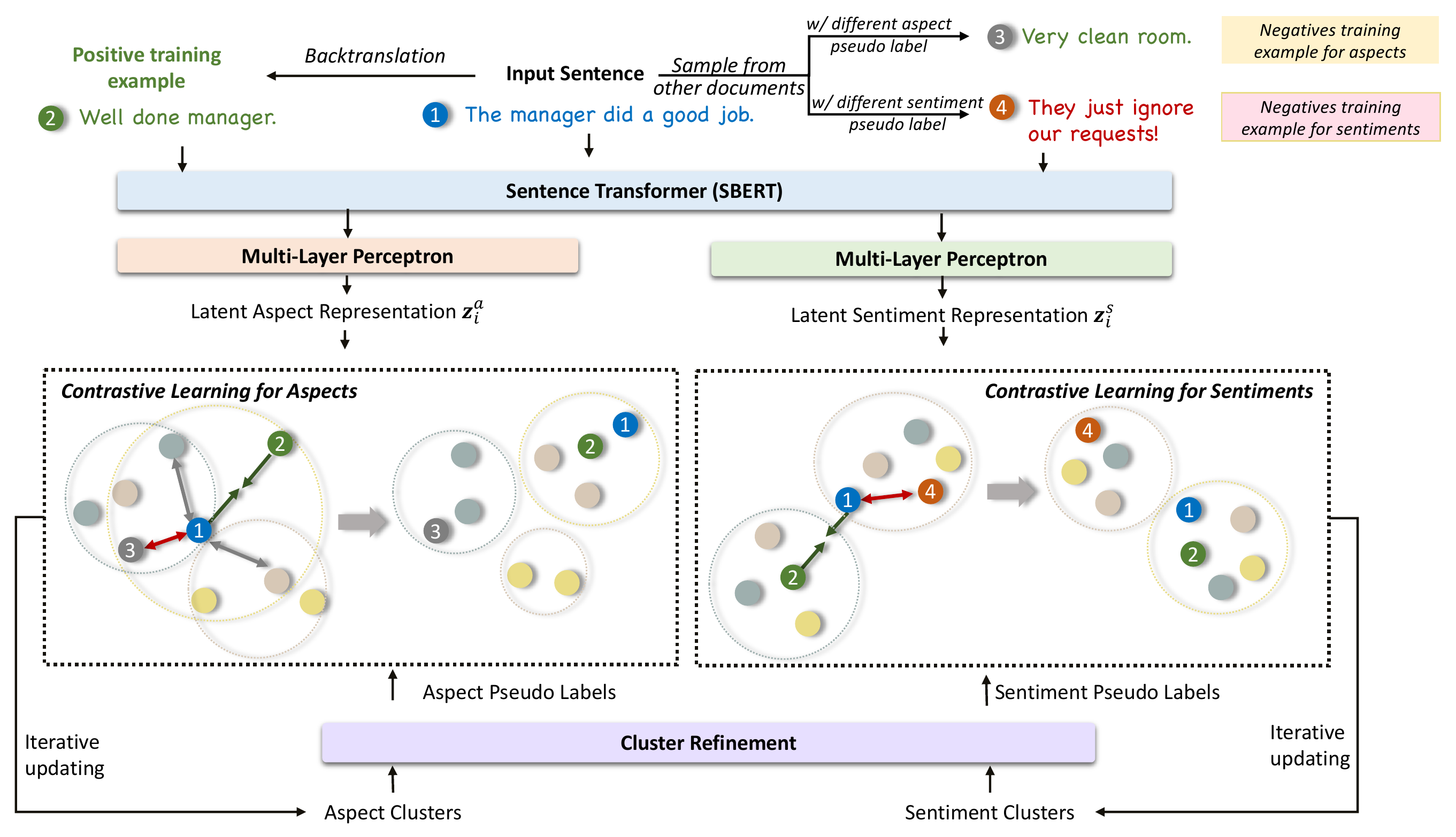}
    \caption{The overall architecture of the Contrastive OpinioN Extraction (\textsc{Cone}) model. Each sentence in a document is encoded using the Sentence Transformer or SBERT, which is then fed to two separate Multi-Layer Perceptrons to generate latent aspect and sentiment representations. The disentanglement of latent aspect and sentiment representations is performed by contrastive learning with positive instances generated by backtranslation and negative instances are selected from different documents with different pseudo labels. The learned latent aspect and sentiment representations are used for generating aspect and sentiment clusters, respectively. The clustering results are used to update the pseudo labels, which are in turn used to adjust the latent aspect and sentiment representations. 
    By iterative updating, the final output contains the keypoints from documents, including the aspects, the related polarities, and the representative sentences. }
    \label{fig:model-structure}
\end{figure*}

\section{Methodology}


Our proposed framework, shown in Figure \ref{fig:model-structure}, consists of three main modules: (1) sentiment and aspect pseudo-labelling; (2) contrastive learning; and (3) clustering refinement. In what follows, we present each module in detail.

\subsection{Sentiment and aspect pseudo-labelling} 

Given a document $d$ consisting of $M_d$ sentences, $d=\{s_1, \cdots, s_{M_d}\}$, we first feed each sentence $s_i$ to a sentence transformer, SBERT, to derive its representation $\bm{e}_i$. We then feed $s_i$ to an off-the-shelf sentiment classifier, the VADER sentiment analyser \cite{vader2014}, to obtain its sentiment pseudo-label $y_i^s$. 
To get the aspect pseudo-label, $y_i^a$, for sentence $s_i$, we perform $K$-means on all sentences in the corpus based on the sentence embeddings generated from SBERT. 
Both sentiment pseudo-labels and aspect clusters will be updated as will be described in the clustering refinement step in Section 3.3. The actual number of aspect clusters $C$ will be determined automatically in the refinement step. The sentence representation $\bm{e}_i$ is further fed into two Multi-Layer Perceptrons (MLPs) to generate the latent aspect representation $\bm{z}_i^a$ and the latent sentiment representation $\bm{z}_i^s$. Note that both sentence representations, $\bm{z}_i^a$ and $\bm{z}_i^s$, and sentiment/aspect pseudo-labels are treated as parameters and will be updated in our framework in an iterative manner.

\subsection{Contrastive Learning}
\label{sec:inference}

In order to extract contrastive opinions from text, we need to map text into disentangled latent factors with one encoding the sentiment information, while another capturing the aspect topics. It has been previously discussed in \cite{locatello2019challenging} that unsupervised learning of disentanglement is impossible without imposing inductive bias on both model and data. But supervised disentangled representation learning is difficult here, since we don't have the access to the true sentiment and aspect labels in our data.  
Instead, we propose to use contrastive learning to update sentence-level sentiment and aspect latent representations in order to disentangle them into separate spaces. The key to contrastive learning is the construction of positive and negative training pairs based on the pseudo-sentiment and pseudo-aspect labels.

\begin{figure*}[htb]
    \centering
    \includegraphics[width=0.55\linewidth]{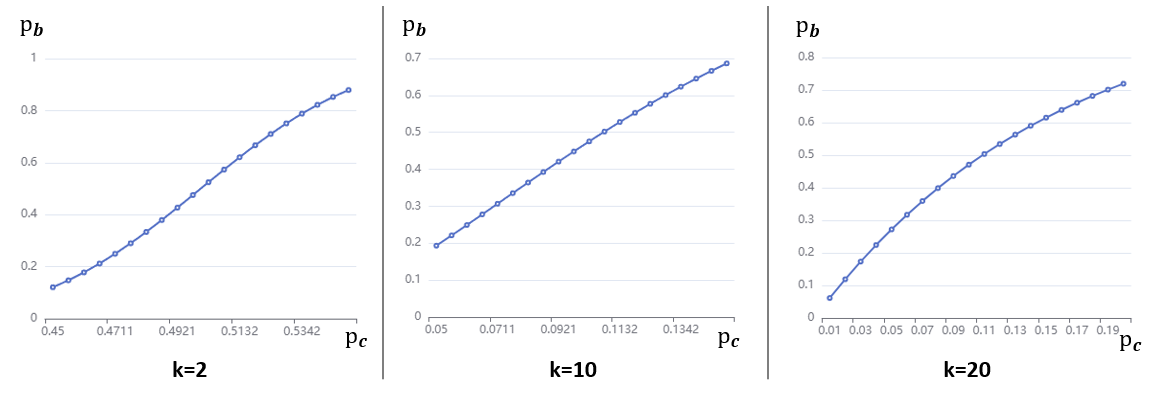}
    \caption{Simulated $p_b$ vs. $p_c$ under various aspect cluster numbers, where $p_b$ is the probability of getting better result using our proposed negative sampling strategy, $p_c$ is the accuracy of aspect clustering.}
    \label{fig:simulated_plots}
\end{figure*}

\paragraph{\textbf{Positive/Negative Instance Selection for Contrastive Learning. }}
For a given sentence $s_{i}$ from document, we further pass the SBERT-encoded output $\bm{e}_{i}$ to two separate Multi-Layer Perceptrons (MLPs) to generate the latent sentiment representation $\bm{z}_{i}^s$ and aspect representation $\bm{z}_{i}^a$. 
To form a positive training pair, for $s_{i}$, we randomly select one sentence $s_{j}$ from the same document with the same pseudo sentiment/aspect label. If none of the other sentences in the same document has the same pseudo label, we generate $s_{j}$ by backtranslation, which is expected to hold the same label as $s_{i}$. 
For negative training instances, we randomly sample sentences from the corpus which are from different documents and with different pseudo labels. 

As we select positive and negative training examples based on pseudo labels, it might be the case that some of our selected positive or negative instances are false positive or false negative examples, which could potentially lead to spurious contrastive learning results. Nevertheless, we can prove that our negative instance sampling strategy is better than random sampling without considering pseudo labels under certain conditions. In particular, let $p_0$ be the proportion of true negatives by randomly sampling negative instances from different documents without considering their associated pseudo labels. 
Let $p_1$ be the proportion of true negatives using our proposed sampling strategy, i.e., sampling negative instances from different documents and with different pseudo labels. 
We have the following bound: 
\begin{theorem}
\label{theorem}
Given $(p_c, k, N)$, where $p_c$ is the accuracy of aspect clustering, k is the number of aspect clusters, N is the sample size such that $N \gg k$, then $p_1 > p_0$ is guaranteed when $p_c > \frac{1}{k}$. 
\end{theorem}

The above theorem\footnote{The proof of \textsc{Theorem} 3.1, \textsc{Lemma} 3.2 are given in Appendix A.} states that if the aspect clustering accuracy is 
better than random guessing, which is $\frac{1}{k}$, then using our proposed negative sampling strategy (i.e., by considering pseudo labels) is guaranteed to be better than random sampling without considering pseudo labels. 
This is plausible since the accuracy of aspect clustering should be better than random guessing in most cases. 
Note that the above theorem is based on the assumption $N \gg k$, which might not be true in practice as the sample size $N$ will be constrained by the training batch size. This brings up the next question of how the model is affected when $N$ is small. 
\begin{lemma}
\label{lemma}
The probability of obtaining better result using our proposed negative sampling strategy compared to random sampling with a given $(p_c, k, N)$ follows
\begin{equation} \small
\begin{aligned}
    p_b & =  f_{B}(i,N,\frac{1}{k})f_{B}(j,i,p_c)F_{B}(\lfloor\frac{(N-i)j}{i}\rfloor,N-i,\frac{1-p_c}{k-1}),
    \label{eq:improvement-main}\\
\end{aligned}
\end{equation}
\end{lemma}
\noindent where $f_B(\cdot)$ is the Probability Mass Function of a Binomial distribution, and $F_B(\cdot)$ is the Cumulative Distribution Function of a Binomial distribution. 
Since there is no closed form of Eq.~(\ref{eq:improvement-main}), we instead plot $p_b$ vs. $p_c$ under different aspect cluster numbers $k$, 
shown in Figure~\ref{fig:simulated_plots}. The results show that $p_c$ needs to be slightly higher than the theoretical lower bound of $\frac{1}{k}$ in order to have a better result compared to random sampling of negative instances without considering psuedo labels. It is also worth noting that when $k=20$, the minimum required aspect clustering accuracy, $p_c\approx 0.11$, which is about the same as the value required when $k=10$. This is much higher than the theoretical bound of $\frac{1}{k}=\frac{1}{20}$. 

\begin{remark}
\label{remark}
Larger batch size can reduce the uncertainty of results, especially with a larger value of aspect cluster number $k$.
\end{remark}

With the decreasing batch size $N$ for training, $\frac{N}{k}$ would drop for a fixed cluster number $k$. That is, the average number of sentences in each cluster becomes smaller. We find in such a case, the probability of correctly filtering false negatives becomes smaller. Since the actual number of aspects is unknown, we can either increase the training batch size $N$ to reduce the uncertainty or increase $p_c$ to improve the clustering performance.

Our analysis above gives the expected aspect clustering accuracy with respect to the aspect cluster number in order to produce reasonable results using our proposed negative instance sampling strategy. We also show that we could make the batch size larger to compensate for the performance drop caused by the larger number of aspect clusters. By doing this, false positive and false negative cases in the sampled training instances can be largely reduced by considering their pseudo labels. 

\paragraph{\textbf{Positive Sample Generation by Data Augmentation for Training Robustness.}} 

We have discussed previously that we can form positive training pairs by selecting sentences bearing the same pseudo label in the same document. However, there is no way to guarantee the correctness of all pseudo labels under unsupervised learning. 
Our preliminary experimental results show that false positives could lead to a mix of related aspects, such as `\emph{Ambience}' and `\emph{Room}'. In addition, it also causes redundant aspect clusters where multiple clusters are about the same aspect. 

To mitigate this problem, we propose to generate positive training pairs
by data augmentation instead of directly selecting sentences from the same document with the same sentiment or aspect pseudo label. 
We explore various synthetic data generation approaches for generating positive training pairs for contrastive learning, including backtranslation \cite{backtranslation} and random masking \cite{randommask}. In backtranslation, a sentence is first translated to another language and then translated back to English. In random masking, a sentence with some tokens randomly masked is fed to a pre-trained language model to predict the missing tokens. The generated sentence and the original sentence will form a positive pair for contrastive learning. As backtranslation gives better results compared to random masking in our preliminary experiments, we use backtranslation in our work here. 

\paragraph{\textbf{Contrastive Learning}}
During training, we randomly sample $N$ sentences in a mini-batch as negative training instances. Together with the positive training instance for each sentence, we will have a total of $2N$ sentences in each mini-batch. Contrastive learning aims to minimise the distance between positive pairs and maximise the distance between negative pairs. 
For a sentence $s_i$, assuming its latent sentiment and aspect representations are $\bm{z}_i^s$ and $\bm{z}_i^a$, and its corresponding positive samples are $\bm{z}_j^s$ and $\bm{z}_j^a$, respectively. The contrastive losses for learning latent sentiment and aspect vectors are defined as:

\begin{equation}\small
    \mathcal{L}_{(i, j)}^s = -\log\frac{\exp(\mbox{sim}(\bm{z}_{i}^s,\bm{z}_{j}^s)/\tau)}{\sum_{k=1}^{2N} \mathbbm{1}_{k\ne i, y_k^s \ne y_i^s, d_k \ne d_i} \exp(\mbox{sim}(\bm{z}_{i}^s,\bm{z}_{k}^s)/\tau)}
\end{equation}
\begin{equation}\small
    \mathcal{L}_{(i, j)}^a = -\log\frac{\exp(\mbox{sim}(\bm{z}_i^a,\bm{z}_j^a)/\tau)}{\sum_{k=1}^{2N} \mathbbm{1}_{k\ne i, y_k^a \ne y_i^a, d_k \ne d_i}\exp( \mbox{sim}(\bm{z}_{i}^a,\bm{z}_{k}^a)/\tau)}
\end{equation}
\begin{equation}\small
    \mathcal{L} = \frac{2}{2N(2N-1)}\sum_{i, j} (\mathcal{L}_{(i, j)}^s + \mathcal{L}_{(i, j)}^a)
\end{equation}
where $N$ is the batch size, $y_i^s$ and $y_i^a$ are the sentiment and aspect pseudo label for sentence $s_i$, respectively, similarly for $y_k^s$ and $y_k^a$, $d_k \ne d_i$ means the sentence $s_k$ and $s_i$ are not in the same document, 
$\tau$ is the temperature controlling the sentence representation concentration in the target space, $\mbox{sim}(\cdot)$ is a similarity measurement function, which is set to the cosine similarity function in our experiments. 

\subsection{Clustering Refinement \& Keypoint Generation}

\paragraph{\textbf{Aspect Clustering Refinement.}} Once the latent aspect representation of each sentence, $\bm{z}_i^a$, is updated in contrastive learning, we next perform clustering on the learned latent aspect vectors. 
After obtaining clustering results, we then refine aspect clusters by merging similar aspect clusters. In order to set an appropriate threshold, we model the distance between sentences, $\delta$, by log-normal distribution:
\begin{equation}\small
    \delta \sim \mathcal{LN}(\mu, \sigma^2), \label{eq:beta-evolution}
\end{equation}
where
\begin{equation}\small
   \mu = \frac{2}{N_S(N_S-1)}\sum_{i, j \neq i}^{N_S} \log(\mbox{sim}(\bm{z}_i^a, \bm{z}_j^a)),\nonumber
\end{equation}
\begin{equation}\small
    \sigma^2 = \frac{2}{N_S(N_S-1)-2}\sum_{i, j \neq i}^{N_S} (\log(\mbox{sim}(\bm{z}_i^a,\bm{z}_j^a))-\mu)^2,\nonumber
\end{equation}
$N_S$ is a sentence subset randomly sampled from all input sentences. The reason for choosing log-normal distribution is because it is a continuous probability distribution of positive variables, which is the case of distance between different sentences. 

We can then estimate the threshold $\alpha$ to merge clusters by percentile $\rho$ of the above log-normal distribution, $\alpha = F_{\delta}(\rho) = P(\delta \leq \rho)$, $\rho$ is a hyper-parameter. The resulting aspect clusters will update the pseudo aspect label of each sentence. The whole process of \emph{contrastive learning} and \emph{aspect clustering refinement} is repeated until the silhouette score of the aspect clusters converges.
\paragraph{\textbf{Sentiment Clustering Update.}} Similarly, we also update the sentiment latent vector of each sentence, $\bm{z}_i^s$, and the pseudo sentiment labels at each iteration. But there is no need to merge clusters as the number of sentiment clusters is fixed across training iterations. 

\paragraph{\textbf{Contrastive Key Point Generation}}

Once a set of aspect clusters is obtained, from which we can output a set of associated sentences ranked by their similarities to their corresponding cluster centroids. To assign a sentiment label to each sentence, we first partition the latent sentiment representations of all sentence in the corpus, $\{\bm{z_{d,i}}^s\}, d=\{1,\cdots,|\mathcal{D}|\}, i=\{1,\cdots,M_d\}$, into three categories, corresponding to the positive, negative and neutral sentiment classes. For each sentence, we can then calculate the similarity between its latent sentiment representation and each of the sentiment cluster centroids to generate the sentiment polarity score.

\section{Experiment}
\paragraph{\textbf{Datasets}}
For contrastive key point extraction, we conduct experiments on the following datasets, the \emph{HotelRec Review} dataset \cite{antognini-faltings:2020:LREC1}, comprises of hotel reviews from TripAdvisor, and the \emph{Vaccine Attitude Detection} (VAD) dataset \cite{naacl2022zhu}, consists of tweets discussing COVID-19 vaccines. These datasets cover two different scenarios that the \emph{HotelRec Review} consists of reviews with multiple sentences, but with a limited number of aspects; while VAD consists of tweets with shorter lengths, but discussing a potentially much larger number of aspects.\footnote{Available aspect-based sentiment analysis datasets, such as \cite{2014SemEval},  are too small 
to fit the purpose of our work which is designed for unsupervised clustering of a large amount of unlabelled data.} 
Each review in the HotelRec Review dataset is accompanied with an overall rating, which is however not used during training. The VAD dataset does not have any sentiment label. Reviews from HotelRec are longer and therefore may discuss multiple aspects. They are first split into sentences before feeding into our framework. Tweets are short due to the character limit and only contain a single aspect in most cases. They are used directly as input to our framework. The dataset statistics are shown in Table \ref{tab:datasets}.

\begin{table}[th]
\centering
\caption{Dataset statistics.}\label{tab:datasets}
\resizebox{0.7\columnwidth}{!}{
\begin{tabular}{lrr}
\hline
\textbf{Dataset}                            & \textbf{HotelRec}         & \textbf{VAD}     \\ \hline
\# Documents                                & 10,000                     & 30,000                \\
\ \ Positive/Neutral/Negative                              & 4k /2k /4k              & -  \\
Ave \#Sents per Doc                               & 5.96                     & -                    \\
Ave \#Words per Sent                                      & 20.22                    & 30.61          \\
\hline
\end{tabular}
}

\end{table}

\paragraph{\textbf{Models for Comparison}} 
We compare with the following baselines, including the start-of-the-art models for key point extraction based on clustering, unsupervised opinion summarisation, contrastive opinion extraction built on LDA. 
We also compare with the Linguistic-Style Transfer model which is not used for contrastive opinion detection originally but for disentanglement of latent factors encoding styles and content:


\noindent \underline{Deep Embedding Clustering (DEC)} \cite{2016:ptdec}, a model based on deep neural networks for simultaneously learning feature embeddings and clustering based on the learned embeddings. 

\noindent \underline{Key Point Analysis(KPA)} \cite{kpa2020}, an unsupervised summarisation framework aims to generate key points, each of which is represented by a set of sentences. The model requires a key point quality classifier trained on some labelled data in order to select high quality key points based on the popularity of the points discussed in text.

\noindent \underline{SemAE} \cite{acl2022chowdhury}, an unsupervised summarisation approach using autoencoder to learn the latent representations, which can be used to generate aspect-specific summaries with pre-defined aspect-denoted seed words. In our experiments, we use the seed words from another hotel dataset called SPACE \cite{2021:qt} for HotelRec and use keywords of aspect definitions from the original paper for the VAD dataset \cite{naacl2022zhu}.

\noindent \underline{ContraVis} \cite{www2019le}, a model based on contrastive topic modelling built on supervised LDA. It is able to generate the document representations, topics and label embeddings for a unified visualisation.


\noindent \underline{Linguistic Style-Transfer Model (LST)}\footnote{LST was not proposed for contrastive opinion extraction. We only use LST for the evaluation of the learned latent sentiment and aspect embeddings.} \cite{2019ACLJohn}, a model combining auxiliary multi-task and adversarial learning objectives to disentangle latent representations of style and content in order to do style transfer. Style labels are required for model training. Here we can treat sentiment as style, while aspect as content. We use the pseudo labels generated by the same pre-trained sentiment classifier used in our experiments as the style labels here. 

\paragraph{\textbf{Parameter Setting}}
We perform sentence tokenisation and remove sentences with only digits or punctuation. For VAD, user mentions are also removed. 
We generate sentiment pseudo labels using  VADER\footnote{\url{https://www.nltk.org/api/nltk.sentiment.vader.html}}.  
\change{For sentence embedding initialisation, we compared the commonly used embedding methods,  BERT\footnote{\url{https://huggingface.co/bert-base-uncased}} and SBERT\footnote{\url{https://huggingface.co/sentence-transformers/all-MiniLM-L6-v2}}, and chose SBERT as it gives better results.
For clustering initialisation, we apply $K$-means\footnote{\url{https://scikit-learn.org/stable/modules/generated/sklearn.cluster.KMeans.html}} with random seeds. The output quality is stable with different seeds.} For hyperparameters, the batch size is set to 128, the initial number of aspect clusters is set to 20, the maximum number of epochs is set to 10, the hyperparameter $\rho$, used for the estimation of threshold for the merge of aspect clusters, is set to 0.05. All models are trained on the whole corpus, while the outputs for HotelRec can be split based on their associated hotels. 

\section{Experimental Results}

\begin{figure*}[htb]
    \centering
    \includegraphics[width=\linewidth]{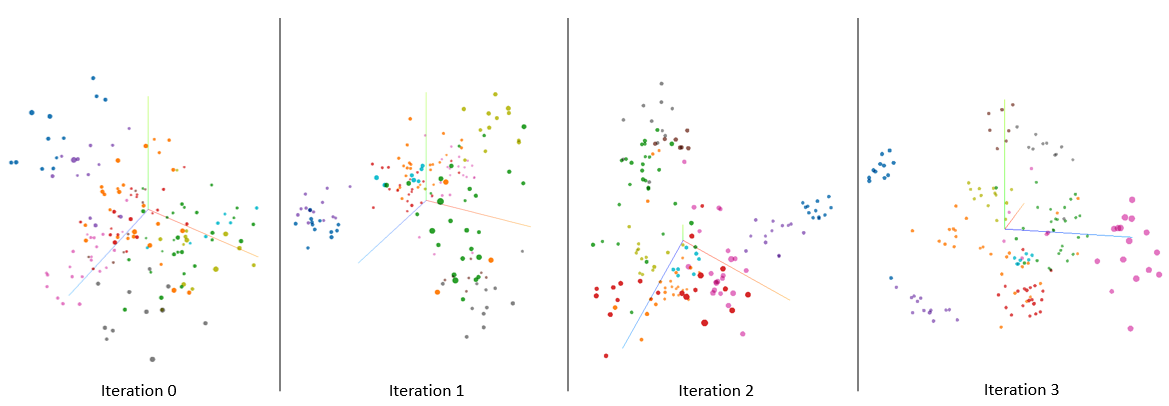}
    \caption{The PCA plots of the \textsc{Cone} latent aspect embeddings on 200 manually annotated sentences from HotelRec, with the aspect embeddings at iteration 0 initialised from SBERT. Each dot represents a sentence, which is colour-coded based on the true aspect label.}
    \label{fig:embedding_pos4aug}
\end{figure*}

\paragraph{\textbf{Key Point Evaluation Results}}
To monitor the update of latent aspect embeddings with the iterative aspect clustering refinement, we randomly selected and labelled 200 sentences from the HotelRec corpus. The PCA plots of aspect embeddings generated from \textsc{Cone} in different training iterations is shown in Figure~\ref{fig:embedding_pos4aug}. We can see that in Iteration 0, the SBERT-initialised aspect embeddings are scatterd in the embedding space without separating into groups. But with the increase of training iterations, we see the updated aspect embeddings exhibiting clearer cluster separations. To quantify the separation of aspect clusters, we use silhouette coefficient \cite{silhouette1987}, $sc(z) = \frac{b(z)-a(z)}{max\{a(z), b(z)\}}$, 
where $a(z)=\frac{\sum_{z' \in C_i, z \neq z'}dist(z,z')}{|C_i|-1}$ is the average distance within the same aspect class, which measures the compactness within the class.
$b(z) = min_{Cj,j \neq i}\frac{\sum_{z' \in C_j}dist(z,z')}{|C_j|}$ is the minimum average distance between aspect embeddings from different classes, which measures the separation of aspect distribution. 
The silhouette coefficient of aspect clusters at different iterations is $0.0287$, $0.1296$, $0.1573$, $0.1941$ from iteration $0$ to $3$. We can observe increased silhouette coefficient values of aspect clusters with the increase of training iterations, indicating improved clustering results.

We next measure the quality of aspect clusters in terms of coherence and lexical diversity. We want to achieve a balance between the semantic coherence and lexical diversity in sentences within each aspect cluster. The aspect coherence, $c = \frac{\sum_{i \neq j}\mbox{sim}(s_i, s_j)}{\sum_{i,j}\mathbbm{1}_{i \neq j}}$, is the average cosine similarity score of embeddings of all possible sentence pairs within a cluster. As different models may use different embeddings, for fair comparison, the sentence embedding here is encoded by SBERT.
 We measure word diversity by following \cite{2016NAACLli} to compute $Div1 = \frac{\#unique\ unigrams}{\#total\ words}$ and $Div2 = \frac{\#unique\ bigrams}{\#total\ words}$.

To measure the overlap across aspect, we calculate the cross aspect distance $d = \frac{\sum_{i \neq j} \mathcal{D}(c_i, c_j)}{\sum_{i,j}\mathbbm{1}_{i \neq j}}$, the average distance of the centroids of all possible aspect cluster pairs, where $c_i$ is the mean of embeddings of all sentences in the aspect cluster $i$.  
We also measure aspect uniqueness in a similar way as topic uniqueness defined in \cite{tmwa2019}. Since the number of clusters varies for different models, we update the original formula by changing the repeated word count across aspect clusters to cosine similarity between aspects, which measures the sentence overlap across aspects. Stopwords are removed when calculating word diversity and word overlap following \cite{2019ACLJohn}.  

\begin{table}[h]
\centering
\caption{The results of Aspect Coherence (\texttt{coh}), Word Diversity (\texttt{div}), which including unigram diversity (\texttt{Div1}) and bigram diversity (\texttt{Div2}), Aspect Distance (\texttt{dis}) and Aspect Uniqueness (\texttt{uni}) calculated for various models. }
\label{tab:output-evaluation}
\resizebox{0.75\columnwidth}{!}{
\begin{tabular}{lccccc}
\toprule
Model                & \multicolumn{3}{c}{Within Cluster}                         & \multicolumn{2}{c}{Between Clusters}          \\ \hline
                     & \multirow{2}{*}{coh} & \multicolumn{2}{c}{div}           & \multirow{2}{*}{uni} & \multirow{2}{*}{dis} \\ \cline{3-4}
                     &                      & Div1       & Div2       &                      &                      \\ \hline
\multicolumn{6}{c}{\textit{HotelRec}}                                                                                         \\ \hline
DEC                    & 0.3952          & 0.8428          & 0.8572          & 1.6706                & 1.3299                \\
KPA                    & \textbf{0.6054} & 0.8321          & 0.8689          & 1.7884                & 1.1406                \\
SemAE                  & 0.4383          & 0.8476          & 0.8499          & 1.2000                & 0.3969                \\
ContraVis              & 0.3842          & 0.8849          & 0.8708          & 1.6533                & 0.7120                \\
\textsc{Cone}                     &0.4792          & \textbf{0.8853} & \textbf{0.8815} & \textbf{2.1797}       & \textbf{1.4052}       \\ \hline
\multicolumn{6}{c}{\textit{VAD}}                                                                                           \\ \hline
DEC                    & 0.5499          & 0.1966          & 0.6087          & 1.0872                & 0.2593                \\
KPA                    & \textbf{0.6836} & 0.2538          & 0.5190          & 1.1497                & 0.9045                \\
SemAE                  & 0.3464          & \textbf{0.2606}          & 0.6214          & 1.0436                & 0.2327                \\
ContraVis              & 0.5860                & 0.2571               & 0.5564                & 1.1154                      & 0.7595                      \\
\textsc{Cone}                     & 0.6187          & 0.2581 & \textbf{0.6257} & \textbf{1.2190}       & \textbf{0.9700}  

\\\bottomrule
\end{tabular}
}
\end{table}

As shown in Table~\ref{tab:output-evaluation}, our model outperforms the other baselines on both datasets across all metrics, except the aspect coherence, where it is inferior to KPA. KPA discards the content that it is less confident with and only outputs aspects on which it has high confidence. This leads to the high coherence value of its generated aspects, but may also result in the information loss due to the discarded content. As will be shown in our human evaluation results in Table~\ref{tab:human-evaluation}, the number of informative clusters of KPA is significantly less than that of \textsc{Cone}.
In addition, as KPA only keeps the best matching sentences, it also discards sentences with the same aspect but different sentiment. This sacrifices the sentiment accuracy of KPA and makes the sentiment split meaningless as one sentiment would be dominating in most cases. Instead, the sentiment accuracy and sentiment split of \textsc{Cone} better reflect the whole corpus and 
making it much easier to compare contrastive opinions. 

\paragraph{\textbf{Disentanglement of Latent Aspect and Sentiment Embeddings}}
\label{sec:embeddings}

To evaluate the disentanglement of latent aspect and sentiment embeddings, we compare the average cosine similarity between aspect and sentiment embeddings for all sentences 
$sim = \frac{\sum_{i}Sim(z^a_i, z^s_i)}{M}$, where $M$ denotes the total number of sentences in a corpus. As baselines other than LST do not learn separate aspect and sentiment embeddings, we only compare our results with LST. 
Intuitively, a lower average cosine similarity between the learned latent aspect embeddings and the sentiment embeddings indicates a better disentanglement result. 
Our model achieves on average the cosine similarity of $0.6355$ on HotelRec and $0.5162$ on VAD,  outperforming LST which gives $0.6881$ on HotelRec and $0.6735$ on VAD, showing a better disentanglement of latent sentiment and aspect embeddings. 

\begin{figure*}[htb]
    \centering
    \includegraphics[width=0.8\linewidth]{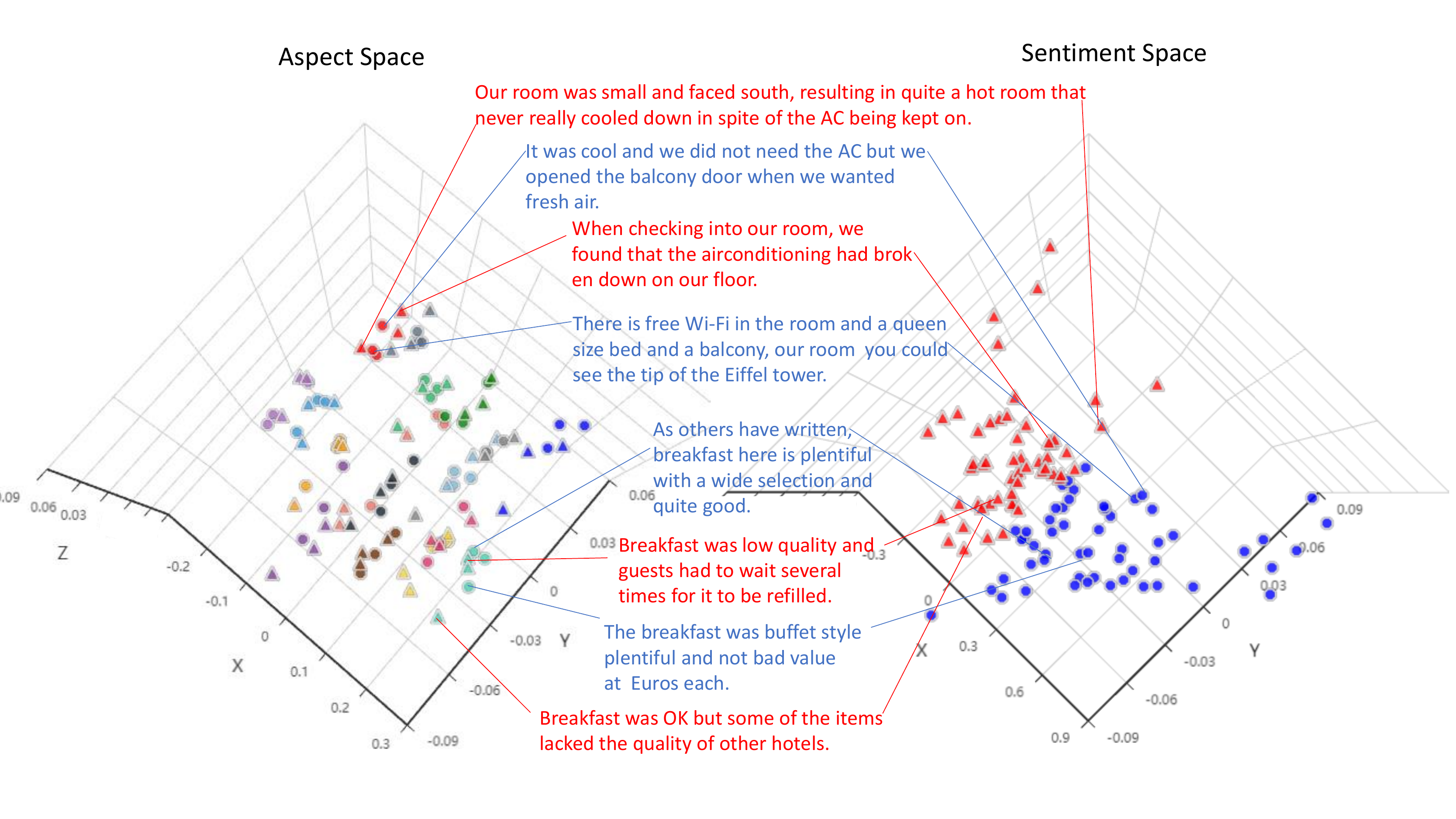}
    \caption{The PCA plots of the latent aspect and sentiment embeddings on HotelRec. Each dot represents a sentence, which is colour-coded based on the \textsc{Cone} outputs. 
    In the aspect space (LHS),  positive tweets are marked by circles, while negative tweets are marked by squares, and different colours indicate different aspects; in the sentiment space (RHS), positive tweets are marked by blue circles and negative tweets by red squares.}
    \label{fig:hotel_aspect}
\end{figure*}

We also show the PCA plot of the learned latent aspect and sentiment embeddings from HotelRec with some representative sentence examples in Figure~\ref{fig:hotel_aspect}. 
It can be observed that sentences are grouped in various aspect clusters. From the example sentences shown in the top aspect cluster, it is clear that the aspect is about the hotel \emph{rooms} with negative comments about the room temperature and positive opinions on the room layout and views. The lower aspect cluster is about \emph{food} with positive opinions centred on wide selections while negative ones complain about the decoration of the breakfast room and food quality.

\paragraph{\textbf{Document-level Sentiment Results}}

\begin{table}[h!]
\centering
\caption{Comparison of document-level classification accuracy, precision, recall and Macro-F1.}
\label{tab:ablation-sentiment}
\resizebox{0.7\columnwidth}{!}{
\small
\begin{tabular}{ccccc}
\toprule
    & Accuracy             & Precision             & Recall             & Macro-F1              \\ \midrule
VADER & 0.5326          & 0.5525          & 0.4962          & 0.4740          \\
\textsc{Cone} & \textbf{0.7137} & \textbf{0.6173} & \textbf{0.6127} & \textbf{0.5854}
\\\bottomrule
\end{tabular}}
\end{table}

\definecolor{blue-green}{rgb}{0.0, 0.87, 0.87}
\definecolor{caribbeangreen}{rgb}{0.0, 0.8, 0.6}
\definecolor{chocolate}{rgb}{0.48, 0.25, 0.0}

\begin{figure}[h!]
    \centering
    \includegraphics[width=0.6\columnwidth]{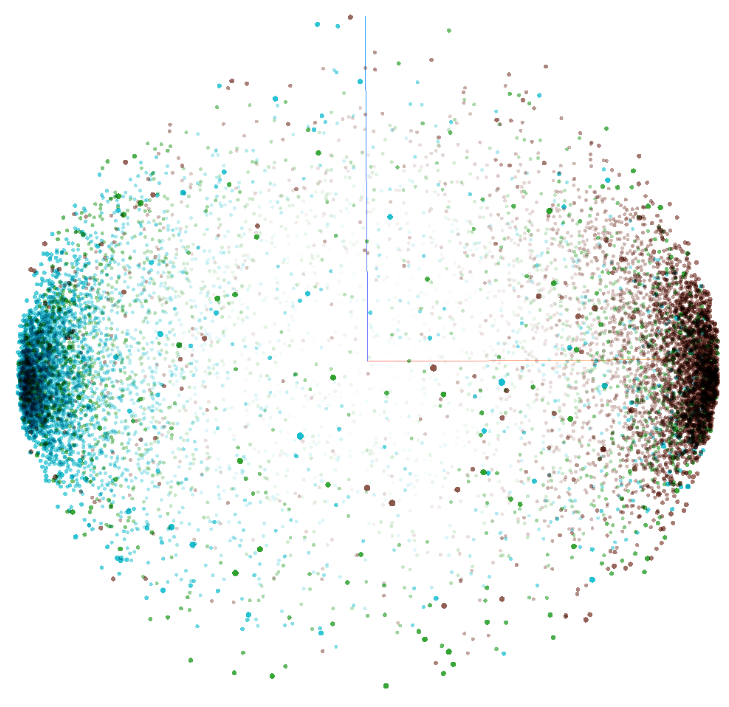}
    \caption{The document-level sentiment PCA plot on HotelRec, where different colours indicate the true labels. (``\emph{\textcolor{blue-green}{positive}}'' in \textcolor{blue-green}{blue}, ``\emph{\textcolor{caribbeangreen}{neutral}}'' in \textcolor{caribbeangreen}{green}, and ``\emph{\textcolor{chocolate}{negative}}'' in \textcolor{chocolate}{chocolate}.)}
    \label{fig:doc-sentiment}
\end{figure}

In our framework, we generate sentence-level pseudo sentiment labels using the VADER sentiment classifier during initialisation. One natural question to ask is how our \textsc{Cone}-generated document-level sentiment labels compare with the sentiment classification results produced by VADER. 
For \textsc{Cone}, we first aggregate the sentence-level sentiment embeddings to derive the document-level sentiment embedding, then calculate the distance between a document embedding, $\bm{e}_d$, and the center of the positive sentiment cluster, denoted as $\bm{c}_p$, and also the centre of the negative sentiment cluster, $\bm{c}_n$, and calculate the sentiment score as 
$s = \frac{|\bm{c}_p-\bm{e}_d|}{|{\bm{c}_p+\bm{c}_n}|}$. We can then classify the document as positive
if $s > 0.1$, negative if $s < -0.1$, and neutral otherwise. The threshold of 0.1 follows the default setting of VADER. From the sentiment classification results shown in Table~\ref{tab:ablation-sentiment}, we can observe that \change{CONE outperforms VADER by a large margin in both accuracy and macro-F1, while being able to extract contrastive opinions, which is not the case with VADER. }

The document-level sentiment embeddings are shown in Figure~\ref{fig:doc-sentiment}, where each document is coloured by its true rating label (positive, neutral and negative). We can see there is clearly a nice separation of positive and negative documents. 

\paragraph{\textbf{Human Evaluation}}

\change{Since our work is designed for unsupervised clustering of a large amount of unlabelled data, there are no sentence-level aspect or sentiment true labels. Therefore, we conduct human evaluation to assess the aspect/sentiment disentanglement and the quality of extracted aspects. 
}
We evaluate the output with the following four measures: (1) \emph{sentiment accuracy}, the proportion of sentences with sentiment labels aligned with the ground-truth sentiment labels. (2) \emph{aspect alignment}, $\frac{\sum_j{max_{i \in C}(n_i)/|C_j|}}{\sum_j\mathbbm{1}}$, which computes the average proportion of sentences discussing the same aspect in a \textsc{Cone}-generated aspect cluster $C_j$. Here, $n_i$ denotes the number of sentences in the human labelled aspect partition $i$. \change{The average number of sentences included in each cluster is also recorded, as there is a trade-off between aspect alignment and the number of sentences included. }
(3) \emph{number of informative aspects}, the number of extracted aspects after removing duplicate aspect clusters. (4) \emph{aspect redundancy rate}, the proportion of redundant aspects in all extracted aspect clusters. \change{Three annotators have been recruited for human evaluation. The inter-annotator agreement score (Fleiss' kappa) is 0.617 for sentiment accuracy, 0.626 for aspect alignment, and 0.866 for aspect redundancy judgement, which shows a good level of agreement among the annotators.}
\begin{table}[h]
\centering
\caption{The human evaluation results in Sentiment Accuracy (\texttt{sent}), Aspect Alignment (\texttt{align}), Average Number of Viewpoints (\#\texttt{viewpoint}), Number of Informative Aspects (\#\texttt{info}) and Aspect Redundancy Rate (\texttt{arr}). }
\label{tab:human-evaluation}
\resizebox{0.95\columnwidth}{!}{
\begin{tabular}{lcccccccc}
\toprule
\multirow{2}{*}{Model}                & \multicolumn{4}{c}{\textit{HotelRec}}                         & \multicolumn{4}{c}{\textit{VAD}}          \\ \cmidrule(lr){2-5} \cmidrule(lr){6-9}
& sent$\uparrow$           & align$\uparrow$            & \#info$\uparrow$     & arr$\downarrow$ & sent$\uparrow$           & align$\uparrow$            & \#info$\uparrow$     & arr$\downarrow$ \\ \hline
DEC       & -              & 0.450          & 10.5          & 0.475    & -              & 0.429          & 16.0           & 0.367            \\
KPA       & 0.618          & \textbf{0.923} & 10.0           & 0.167 & 0.608          & \textbf{0.859}          & 8.3           & 0.167  \\
SemAE     & -              & 0.583          & 7.0           & 0.125    & -              & 0.674          & 7.0           & \textbf{0.125}         \\
ContraVis & -              & 0.498          & 6.3           & 0.424          & -              & 0.645          & 15.7          & 0.175    \\
CONE      & \textbf{0.715} & 0.835          & \textbf{15.3} & \textbf{0.098}  & \textbf{0.695} & 0.782 & \textbf{16.7} & 0.167
\\\bottomrule
\end{tabular}
}
\end{table}

\change{From the results shown in Table~\ref{tab:human-evaluation}, we can observe that for sentiment accuracy, only KPA can generate sentence-level sentiment labels among the baselines. But it shows inferior results compared to \textsc{Cone}. We find \textsc{Cone} performs better in detecting the sentiment of ironic sentences such as ``\emph{If I were caught in the middle of a blizzard, I'd probably consider staying here.}'' and sentences without explicit sentiment expressions 
such as ``\emph{Kindly think about it if someone had the possibility of access to our room whilst we weren't inside.}''. 
For aspect alignment, both KAP and \textsc{Cone} clearly outperform all the other models, with KPA better than \textsc{Cone}. However, 
KPA discarded lower-confidence key points and their supporting evidence, resulting in 
a lower number of informative aspects discovered.
On the contrary, \textsc{Cone} produces the largest number of  informative aspect clusters with a very low aspect redundancy rate compared to the baselines on HotelRec. On the Twitter VAD corpus, \textsc{Cone} has the same aspect redundancy rate as KPA, but extracts the informative aspect clusters double the size of that of KPA. Overall, \textsc{Cone} is able to identify more informative aspect clusters which contain less  conflicting viewpoints, while achieving relatively higher aspect coherence and lower aspect redundancy.}

\paragraph{\textbf{Ablation Studies}}

We investigate the contribution of various components by conducting ablation studies and show the results in Table \ref{tab:ablation-aspect}. 
\change{For the construction of positive samples in contrastive learning, we have explored replacing backtranslation (as in original \textsc{Cone} with (1) randomly choosing sentences from the same document with the same pseudo-label (\texttt{-BT+Doc}); or (2) using pre-trained language models to generate positive training instances by randomly masking some words in an original sentence (\texttt{-BT+RM}). For the construction of negative samples in contrastive learning, we have explored randomly choosing sentences from different documents without considering their pseudo labels, i.e., without the denoising strategy (\texttt{-denoise}). We have also experimented with completely removing the contrastive learning step (\texttt{-ContrL}). 
For clustering, we have explored (1) only performing a single K-means step without clustering refinement (\texttt{-ClusterRef}); or (2) Clustering with a Gaussian mixture model (\texttt{-kmeans+GMM}). } 
For batch size in contrastive learning, we have compared the results with the batch size of 32, 64, 128.

\begin{table}[h]
\centering
\caption{Results with removed component \& different batch sizes: replace backtranslation  with randomly choosing positive training instances from the same document with the same pseudo-label (\texttt{-BT+Doc}), replace backtranslation with random masking (\texttt{-BT+RM}), remove the denoising strategy in contrastive learning (\texttt{-denoise}), remove contrastive learning (\texttt{-ContrL}), remove clustering refinement (\texttt{-ClusterRef}), replace $k$-means with GMM (\texttt{-kmeans+GMM}). Results of \textsc{Cone} with batch size of 32, 64, 128 are also compared here.}
\label{tab:ablation-aspect}
\resizebox{0.95\columnwidth}{!}{

\begin{tabular}{lccccl}
\toprule
Model                & \multicolumn{3}{c}{Within Cluster}                         & \multicolumn{2}{c}{Between Clusters}          \\ \hline
                     & \multirow{2}{*}{coh} & \multicolumn{2}{c}{div}           & \multirow{2}{*}{uni} & \multirow{2}{*}{dis} \\ \cline{3-4}
                     &                      & Div1       & Div2       &                      &                      \\ \hline
\multicolumn{6}{c}{\textit{HotelRec}}                                                                                         \\ \hline
\textsc{Cone}-BT+Doc          & 0.4182          & \textbf{0.9488}          & 0.8905          & 1.7872                & 1.1858                \\    
\textsc{Cone}-BT+RM         & 0.4361          & 0.8635          & 0.8791         & 2.0497                & 1.3239                \\ 
\textsc{Cone}-denoise          & 0.4094          & 0.9480         & \textbf{0.8907}          & 1.7672                & 1.2382                \\
\textsc{Cone}-ContrL        &0.4582          & 0.8748          & 0.8845          & 2.0548                & 1.2181                \\
\textsc{Cone}-ClusterRef         & 0.4386          & 0.8541          & 0.8727          & 2.0639                & 1.3058                \\
\textsc{Cone}-kmeans+GMM                  & 0.4418          & 0.8711 & 0.8891 & 1.9055      & 1.3976       \\
\textsc{Cone}(32)                  & 0.4124          & 0.8880 & 0.8807 & 1.7603       & 1.2389       \\ 
\textsc{Cone}(64)                  & 0.4440          & 0.8881 & 0.8806 & 1.8982       & 1.2615       \\ 
\textsc{Cone}(128)                  &\textbf{ 0.4792}          & 0.8853 & 0.8815 & \textbf{2.1797}       & \textbf{1.4052}       \\ \hline
\multicolumn{6}{c}{\textit{VAD}}                                                                                \\ \hline
\textsc{Cone}-BT+RM              & 0.5859          & 0.2478          & 0.5661          & 1.1818                & 0.9142                \\  
\textsc{Cone}-denoise          & 0.5972          & 0.2018          & \textbf{0.6351}          & 1.1147                & 0.7377                \\
\textsc{Cone}-ContrL             &  0.6059         & 0.1399          & 0.3170          & 1.1224                & 0.7285                \\
\textsc{Cone}-ClusterRef         & \textbf{0.6245} & 0.2573          & 0.5571          & 1.1677                & 0.9045                \\
\textsc{Cone}-kmeans+GMM                    & 0.5947          & 0.2519 & 0.6022 & \textbf{1.2635}       & 0.8116 \\
\textsc{Cone}(32)                  & 0.6034          & 0.2042 & 0.6300 & 1.1196       & 0.8013       \\ 
\textsc{Cone}(64)                  & 0.6123          & 0.2120 & 0.6261 & 1.1229       & 0.8888       \\ 
\textsc{Cone}(128)                   & 0.6187          & \textbf{0.2581} & 0.6257 & 1.2190       & \textbf{0.9700}  
\\\bottomrule
\end{tabular}
}
\end{table}

In general, we observe a consistent drop in all evaluation metrics using the ablated variants. Replacing backtranslation by randomly selecting positive instances from the same document with the same pseudo-label  results in noticeable coherence and distance drop. Replacing backtranslation by random masking for positive instance generation leads to coherence reduction. Removing contrastive learning is essentially to perform clustering on sentences only, which leads to worse performance on all metrics. 
Selecting negative instances without considering pseudo labels in contrastive learning leads to significant reduction in the coherence and distance between aspect clusters. Removing aspect clustering refinement improves coherence on VAD, but suffers a large drop on all other metrics. 
Using $K$-means generally gives better results compared with using GMM for clustering, especially in coherence. For different sizes of training batches, our experimental results agree with the theory shown in Lemma~\ref{lemma} that a larger training batch size generally leads to better performance. The best performance is achieved using the batch size of 128, outperforming the other two in most evaluation metrics.

\section{Conclusion}
We introduce \textsc{Cone}, the Contrastive OpinioN Extraction model, which is able to disentangle the latent aspect and sentiment representations of sentences and automatically extract contrastive opinions despite using no manual label information and aspect-denoted seed words. \change{It can identify more informative aspects, while keeping the conflicting viewpoints and redundant aspects low. In addition, it can quantify the relative popularity of aspects and their associated sentiment distributions for easy user inspection.} 
Experimental results based on the hotel reviews from \emph{HotelRec} and tweets from \emph{VAD} demonstrate the effectiveness of \textsc{Cone} for unsupervised opinion extraction. 

\begin{acks}
This work was supported in part by the UK Engineering and Physical Sciences Research Council (grant no. EP/T017112/2, EP/V048597/1, EP/X019063/1). YH is supported by a Turing AI Fellowship funded by the UK Research and Innovation (grant no. EP/V020579/2). 
\end{acks}


\appendix

\setcounter{table}{0}
\renewcommand{\thetable}{A\arabic{table}}
\setcounter{figure}{0}
\renewcommand{\thefigure}{A\arabic{figure}}
\setcounter{equation}{0}
\renewcommand{\theequation}{A\arabic{equation}}

\section{Full proof} \label{appendix:proof} 
\begin{proof}(for Theorem~\ref{theorem})
Let $n_i(C)$ be  the number of sentences satisfying condition $C$ in the training batch, the proportion of false negatives $p_n$ can be defined as:
\begin{equation}
    p_n = \frac{\mathbb{E}(n({y_i \neq y_j})) - \mathbb{E}(n({y'_i = y'_j | y_i = y_j}))} {N-\mathbb{E}(n({y'_i = y'_j | y_i = y_j}))-\mathbb{E}(n({y'_i = y'_j | y_i \neq y_j}))},
\end{equation}

\noindent where $y_i \in Y$ is the ground truth label and $y_i' \in Y'$ is the pseudo label of the corresponding sentence $s_i$. $\mathbb{E}(\cdot)$ denotes expectation. $n(\cdot)$ and $n(\cdot|\cdot)$ are the counting functions without or with a given condition respectively. In order to guarantee $p_n \leq \frac{1}{k}$, which yields
\begin{equation}
\begin{aligned}
    p_n = &\frac{\frac{N}{k} - \frac{Np_c}{k}}{N - \frac{N(1-p_c)}{k} - \frac{Np_c}{k}} \leq \frac{1}{k}
\label{eq:mask_bound}
\end{aligned}
\end{equation}
\noindent we need to have $p_c \geq \frac{1}{k}$, where $p_c$ is the accuracy of aspect clustering, $N$ is the sample size, and $k$ is the number of aspect clusters. 
\end{proof}

\begin{proof} (for Lemma~\ref{lemma})
\begin{equation} \small
\begin{aligned}
    p_b & = \sum^N_{i=0}\sum^i_{j=0}\sum^{N-i}_{k=0}\mathbb{P}(N_f=i,N_m=j,N_i=k)\mathbbm{1}_{improved} \\
    & = \mathbb{P}(N_f=i)\mathbb{P}(N_m=j|N_f=i)\mathbb{P}(\frac{i-j}{N-j-k}< \frac{i}{N}|N_f=i, N_m=j) \\
    & = f_{B}(i,N,\frac{1}{k})f_{B}(j,i,p_c)F_{B}(\lfloor\frac{(N-i)j}{i}\rfloor,N-i,\frac{1-p_c}{k-1})
    \label{eq:improvement}
\end{aligned}
\end{equation}
$N_f$ is the number of false negative instances, $N_m$ is the number of correctly filtered positive instances, 
$N_i$ is the number of incorrectly filtered negative instances. 
\end{proof}

\bibliographystyle{ACM-Reference-Format}
\bibliography{sample-base}

\end{document}